\documentclass[sigconf]{acmart}

\usepackage{booktabs} 
\usepackage{subfigure}
\usepackage{colortbl}
\usepackage{verbatim}
\usepackage{bm}
\usepackage{soul}
\usepackage{bbm}
\usepackage{bigstrut,bigdelim}
\usepackage{paralist}

\usepackage{algorithm}
\usepackage{algorithmic}
\usepackage{makecell}
\usepackage{diagbox}
\usepackage{multirow}
\usepackage{float}
\usepackage{mathrsfs}
\newcommand{\thickhline}{\noalign{\hrule height 1pt}}

\setcopyright{acmlicensed}

\acmConference[CIKM'20]{The 29th ACM International Conference on Information and Knowledge Management} 
\acmYear{2020}
\copyrightyear{2020}

\newcommand{\zy}[1]{\textcolor{brown}{\textbf{[ZY: #1]}}}

\copyrightyear{2020}
\acmYear{2020}
\setcopyright{acmlicensed}\acmConference[CIKM '20]{Proceedings of the 29th ACM International Conference on Information and Knowledge Management}{October 19--23, 2020}{Virtual Event, Ireland}
\acmBooktitle{Proceedings of the 29th ACM International Conference on Information and Knowledge Management (CIKM '20), October 19--23, 2020, Virtual Event, Ireland}
\acmPrice{15.00}
\acmDOI{10.1145/3340531.3411967}
\acmISBN{978-1-4503-6859-9/20/10}

\begin{document}
\fancyhead{}
\title{Learning to Detect Relevant Contexts and Knowledge for \\Response Selection in Retrieval-based Dialogue Systems}




\author{Kai Hua}
\authornote{Equal contribution.}

\affiliation{%
  \institution{Computer Center, Peking University}
}
\email{kifish@pku.edu.cn}

\author{Zhiyuan Feng}
\authornotemark[1]

\affiliation{%
  \institution{Department of Computer Science and Technology, Peking University}
}
\email{fengzhiyuan@pku.edu.cn}

\author{Chongyang Tao}
\affiliation{%
   \institution{ Wangxuan Institute of Computer Technology , Peking University}
}
\email{chongyangtao@pku.edu.cn}

\author{Rui Yan}
\authornote{Corresponding authors: Rui Yan (ruiyan@pku.edu.cn) and Lu Zhang (zhanglucs@pku.edu.cn) .}
\affiliation{%
  \institution{Wangxuan Institute of Computer Technology, Peking University \\Beijing Academy of Artificial Intelligence (BAAI)}
}
\email{ruiyan@pku.edu.cn}

\author{Lu Zhang}
\authornotemark[2]
\affiliation{%
  \institution{Department of Computer Science and Technology, Peking University}
}
\email{zhanglucs@pku.edu.cn}

%
%
\ccsdesc[500]{Information systems~Retrieval models and ranking}

\begin{abstract}
Recently, knowledge-grounded conversations in the open domain gain great attention from researchers. Existing works on retrieval-based dialogue systems have paid tremendous efforts to utilize neural networks to build a matching model, where all of the context and knowledge contents are used to match the response candidate with various representation methods. 
Actually, different parts of the context and knowledge are differentially important for recognizing the proper response candidate, as many utterances are useless due to the topic shift. Those excessive useless information in the context and knowledge can influence the matching process and leads to inferior performance. To address this problem, we propose a multi-turn \textbf{R}esponse \textbf{S}election \textbf{M}odel that can \textbf{D}etect the relevant parts of the \textbf{C}ontext and \textbf{K}nowledge collection (\textbf{RSM-DCK}). Our model first uses the recent context as a query to pre-select relevant parts of the context and knowledge collection at the word-level and utterance-level semantics. Further, the response candidate interacts with the selected context and knowledge collection respectively. In the end,  
The fused representation of the context and response candidate is utilized to post-select the relevant parts of the knowledge collection more confidently for matching.
We test our proposed model on two benchmark datasets. Evaluation results indicate that our model achieves better performance than the existing methods, and can effectively detect the relevant context and knowledge for response selection. 
\end{abstract}

\settopmatter{printacmref}

\keywords{deep neural network, matching, multi-turn response selection, \\retrieval-based conversation, knowledge-grounded conversation}

\maketitle

\section{Introduction} \label{sec:introduction}

\begin{table*}[t!]
\centering
\caption{An example of document-grounded dialogue from Persona-Chat dataset.}
\resizebox{0.64\textwidth}{!}{
\begin{tabular}{c|l}
\thickhline
\multirow{4}{*}{\textbf{A's profile}} & {horror movies are my favorites.} \\ 
& {i am a stay at home dad.}   \\ 
& {my father used to work for home depot.} \\
& {i spent a decade working in the human services field.}  \\
& {i have a son who is in junior high school.} \\
\hline
\multirow{4}{*}{\textbf{B's profile}} & i read twenty books a year.  \\
& i am a stunt double as my second job.                               \\
& i only eat kosher.                       \\
& i was raised in a single parent household. \\ \hline
\multirow{7}{*}{\textbf{Context}} & {\textbf{A}: hello what are doing today?} \\
& {\textbf{B}: i am good, i just got off work and tired, i have two jobs.}             \\
& {\textbf{A}: i just got done watching a horror movie} \\
& {\textbf{B}: i rather read, i have read about 20 books this year.}             \\
& {\textbf{A}: wow! i do love a good horror movie. loving this cooler weather} \\

& {\textbf{B}: but a good movie is always good.}             \\
& {\textbf{A}: yes! my son is in junior high and i just started letting him watch them too }\\
\hline
\textbf{True response} & {i work in the movies as well.} \\
\hline
\textbf{False response} & {that is great! are you going to college ?}   \\ 
\thickhline
\end{tabular}
}
\label{tab:example}
\end{table*}

The human-machine conversation is the ultimate goal of artificial intelligence. Recently, building a conversation system with intelligence has drawn increasing interest in academia and industry. Existing studies can be generally categorized into two groups.
The first group is retrieval-based dialogue systems~\cite{wu2017sequential,yan2016learning,zhou2018multi,tao-etal-2019-one,henderson-etal-2019-training} which select the proper response from the response candidates under the given user input or dialogue context, and have been applied in many industrial products such as XiaoIce from Microsoft \cite{shum2018eliza} and AliMe Assist from Alibaba \cite{li2017alime}. The second group is generation-based dialogue systems \cite{serban2015building,li2015diversity,shangL2015neural} which generate the response word by word under an encoder-decoder framework \cite{serban2015building,shangL2015neural}. In this paper, we focus on response selection tasks in the retrieval-based dialogues.

The key to the traditional response selection task is to measure the matching degree between the dialogue context and response candidate. Early studies focus on the single-turn dialogue systems with the user input as the query \cite{wang2013dataset, hu2014convolutional,wang2015syntax}, and recently shift to the multi-turn dialogue systems with both the dialogue context and the user input considered, aiming at ranking response candidates by calculating semantic relevance between the dialogue context and response candidates. The sequential matching network \cite{wu2017sequentialframe} proposes a representation-matching-aggregation framework which is followed by subsequent work. The model conducts matching between utterances and the response candidate on the word and utterance level, and aggregates matching features to obtain the matching score. The deep attention matching network \cite{zhou2018multi} captures matching features on multiple levels of granularity. Interaction over interaction network \cite{tao-etal-2019-one} aggregates matching features and supervises those at each level of granularity directly.

Human conversations tend to be related to background knowledge. For instance, when two persons talk about a movie, information about the movie is already some parts of the prior knowledge in their brains. Based on this observation, it is necessary for chatbots to be grounded in external knowledge, which can make human-machine conversations more practical and engaging~\cite{dinan2018wizard}. Therefore, the highlight of research in the dialogue systems has shifted to incorporating the external knowledge into chatbots recently. The Starspace network \cite{wu2018starspace} learns the task-specific embedding and selects the appropriate response by the cosine similarity between the dialogue context concatenated by the associated document and the response candidate. The profile memory network \cite{zhang2018personalizing} uses the dialogue context as the query and performs attention over the document to combine with the query, and then measures the similarity between the fused query and response candidate.
The document grounded matching network \cite{ijcai2019-756} makes the dialogue context and document interact with each other and interact via the response candidate individually later with the hierarchical attention mechanism.
{Dual-interaction-matching-network \cite{gu-etal-2019-dually} makes the dialogue context and document interact with the response candidate respectively via the cross-attention mechanism.}

Existing work on retrieval-based dialogue systems has paid tremendous efforts to utilize the neural network to build various text-matching model, in which all of the context and knowledge contents are used to match the response candidate. Actually, some parts of the context and knowledge collection are irrelevant to the response candidate, which may introduce some noise confusing the model, thus leading to inferior performance. To illustrate the problem, we present an example in table \ref{tab:example}. In the recent dialogue context, the current topic is about movies, so the previous utterances about movies should be given more weight, and utterances that are unrelated to movies (e.g. the fourth utterance of the dialogue context) should be given less weight. Similarly, 
in the knowledge collection, for instance, the second sentence of the B's profile should be assigned more weight than other profile sentences.
This example indicates that different parts of the context and knowledge are differentially important for recognizing the proper response candidate, as many utterances are useless. Those excessive useless information in the context and knowledge can influence the matching process.

To address the above problem, we propose a multi-turn \textbf{R}esponse \textbf{S}election \textbf{M}odel
that can \textbf{D}etect the relevant parts of the \textbf{C}ontext and \textbf{K}nowledge collection (\textbf{RSM-DCK}). For simplicity of expression, we refer to the knowledge collection as documents, however, RSM-DCK can be adapted to other types of knowledge resources such as knowledge graphs. The model first uses the recent utterance of the dialogue context as a query to pre-select relevant parts of the dialogue context and document. Then the selected dialogue context and document interact with the response candidate individually by cross-attention, 
and an BiLSTM \cite{hochreiter1997long, graves2013speech} is employed to aggregate the matching features of the context, document, and response candidate respectively. Due to the inter-dependency and temporal relationship among utterances in the dialogue context, another BiLSTM is adopted to accumulate the dialogue context. 
To select the most relevant sentences in the knowledge collection, the fused representation of the pre-selected dialogue context and response candidate is utilized as the query to post-select the document with the attention mechanism for the reason that sentences in the document are relatively independent and the true response tends to be solely related to one of them.

\begin{figure*}[t!]
    \centering
    \includegraphics[width=0.85\linewidth]{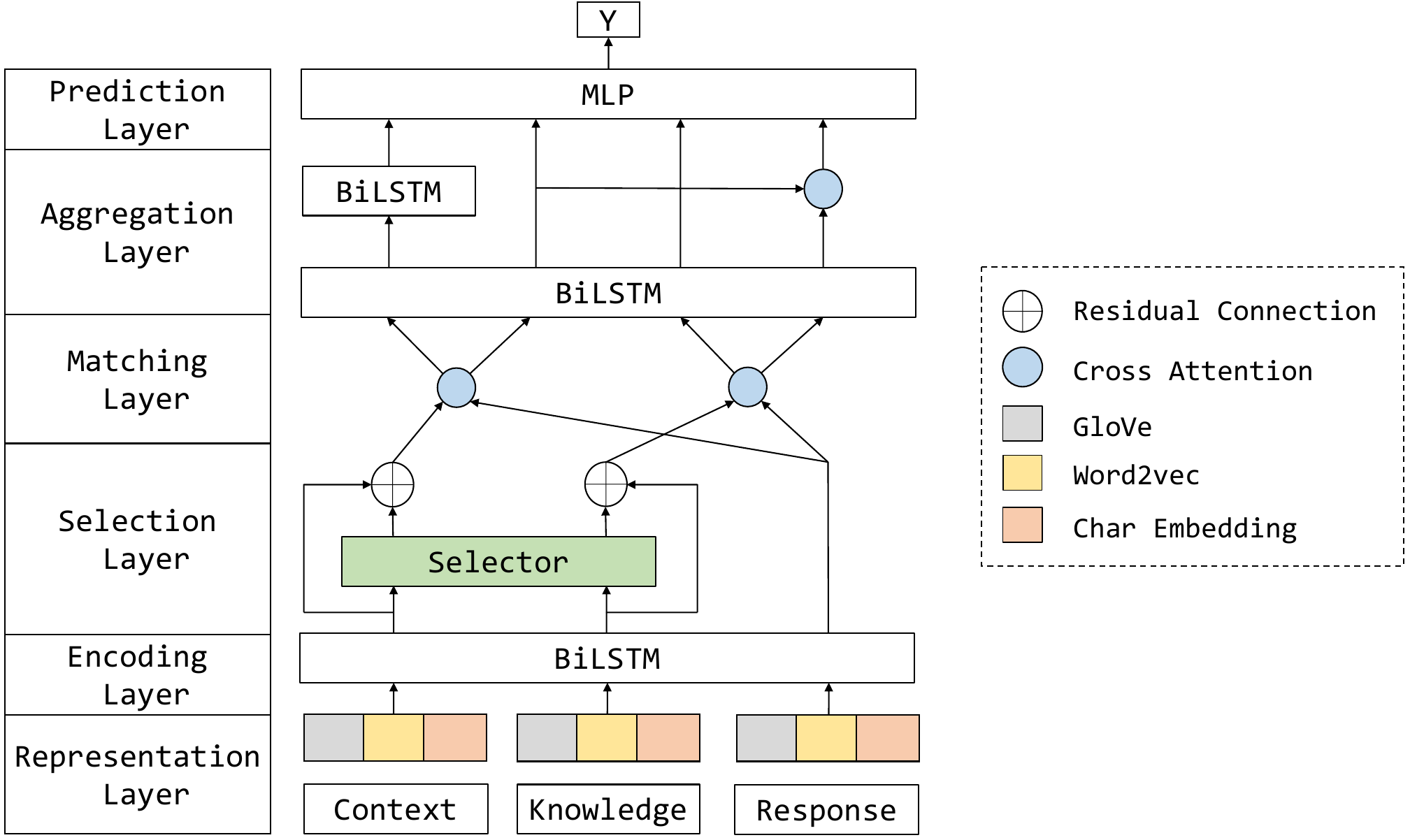}
    \caption{Architecture of the proposed RSM-DCK. }
    \label{fig:model_overview}
\end{figure*}

We conduct experiments on two benchmark datasets for multi-turn knowledge-grounded response selection: the Original/Revised Persona-Chat dataset \cite{zhang2018personalizing}, and the CMUDoG dataset \cite{cmu_dog_emnlp18}. The evaluation results indicate that our model outperforms previous models in terms of all metrics on all the datasets. Compared with dual-interaction-matching-network, the best performing baseline on all the two benchmarks, our model achieves $0.9\%$ absolute improvement on $R_{20}@1$ on the Original Persona-Chat dataset, $1.2\%$ absolute improvement on $R_{20}@1$ on the Revised Persona-Chat dataset, and $0.7\%$ absolute improvement on $R_{20}@1$ on the CMUDoG dataset.

\section{Related Work}
Retrieval-based open-domain dialogue systems learn a matching model to calculate the similarity between the user input and the candidate response for response selection. Early studies focus on the single-turn dialogue with the user input regarded as the query \cite{wang2013dataset,hu2014convolutional,wang2015syntax}, and recently move to the multi-turn dialogue where the dialogue history and the user input are both taken into account. Representative methods include the Dual LSTM \cite{lowe2015ubuntu}, the deep learning-to-respond architecture \cite{yan2016learning}, the multi-view matching model \cite{zhou2016multi}, the sequential matching network \cite{wu2017sequentialframe}, the deep attention matching network \cite{zhou2018multi}, the multi-representation fusion network \cite{tao2019multi}, the interaction-over-interaction matching network \cite{tao-etal-2019-one}, and the interaction matching network \cite{gu2019interactive}.

Knowledge is crucial to the dialogue in the real world. The lack of knowledge makes dialogue systems suffer from semantics issues \cite{zhou2018commonsense}, consistency issues \cite{li2016persona,ijcai2018-595,DBLP:journals/corr/abs-1901-09672}, and interactiveness issues \cite{zhou2017emotional,rao-daume-iii-2018-learning,wang-etal-2018-learning-ask}. Recently, incorporating knowledge into the dialogue has been shown beneficial \cite{ijcai2019-756,gu-etal-2019-dually}. Existing work for knowledge-grounded dialogue systems can be generally categorized into two groups by the type of knowledge. 
The first group is document-grounded dialogue systems where the documents are unstructured texts such as user profiles and Wikipedia articles \cite{zhang2018personalizing,cmu_dog_emnlp18,dinan2018wizard,ijcai2019-756,gu-etal-2019-dually}. The second group is knowledge-graph-grounded dialogue systems where the knowledge graph is a large network of entities which contains semantic types, properties, and relationships between entities
\cite{zhou2018commonsense,DBLP:conf/aaai/YoungCCZBH18,liu-etal-2019-knowledge}.
However, either of the groups can be influenced by irrelevant parts of the context and knowledge collection, thus leading to a severe performance drop.

Recently, Multi-turn dialogue context selection has been widely explored in the response generation and response selection. In the response generation, Xing et al. \cite{DBLP:journals/corr/XingWWZHM17} propose a hierarchical recurrent attention network where the hidden state in the decoder is utilized to select important parts of the dialogue context on both the word- and utterance-level via the hierarchical attention mechanism. 
To effectively model the context of a dialogue, Zhang et al. \cite{zhang-etal-2018-context} propose two types of attention mechanisms (including dynamic and static attention) to weigh the importance of each
utterance in a conversation and obtain the contextual representation.
Zhang et al. \cite{zhang-etal-2019-recosa} propose a model named ReCoSa where the self-attention mechanism is employed to update both the context and masked response representation, and the attention weights between each context and response representations are computed as the relevant score to guide the decoding process.
In the response selection, Zhang et al. \cite{zhang2018dua} propose a deep utterance aggregation model where the last utterance representation refines the preceding utterances to obtain the turns-aware representation and then the self-attention mechanism is applied among utterances. Yuan et al. \cite{yuan-etal-2019-multi} propose a multi-hop selector network where the latter parts of the dialogue context are used as the query to select the relevant utterances on both the word- and utterance-level.
As the knowledge sometimes contains a lot of redundant entries in knowledge-grounded conversation, Lian et at. \cite{DBLP:journals/corr/abs-1902-04911} propose a model with the knowledge selection mechanism which leverages both prior and posterior distributions over the knowledge to facilitate knowledge selection. Kim et al. \cite{Kim2020Sequential} propose a sequential knowledge transformer that employs a sequential latent variable model to better leverage the response information for the proper choice of the knowledge collection in multi-turn dialogue. 
Different from previous work, our model both selects the dialogue context and knowledge collection with the carefully designed selection mechanism.

\section{Problem Formalization}
\label{sec:length}
Suppose that we have a dialogue dataset with knowledge collection $\mathcal {D}=\{c_i,k_i,r_i,y_i\}^{N}_{i=1}$, where $c_i = \{u_{i,1},u_{i,2},...,u_{i,m_i} \}$  represents a conversational context and $m_i$ is the number of utterances in the conversational context; $k_i = \{k_{i,1},k_{i,2},...,k_{i,n_i} \}$ represents a collection of knowledge and $n_i$ is the number of sentences in the collection of knowledge; $r_i$ denotes a candidate response; and $y_i \in \{0,1\}$ is the label. $y_i = 1 $ indicates that $r_i$ is a proper response for $c_i$ and $k_i$, otherwise, $y_i=0$. $N$ is the number of training samples. The task is to learn a matching model $g(c,k,r)$ from $\mathcal{D}$, and thus for a new context-knowledge-response triple $(c,k,r)$, $g(c,k,r)$ returns the matching degree between $r$ and $(c,k)$.

\section{Model}

\subsection{Model Overview}
We propose a matching network with deep interaction to model $g(c,k,r)$. Figure \ref{fig:model_overview} shows the architecture of RSM-DCK,
which consists of six layers, namely the representation layer, encoding layer, selection layer, matching layer, aggregation layer, and predication layer. In the representation layer, multiple types of embedding are adopted to represent words. In the encoding layer, an BiLSTM is adopted to encode the context, knowledge collection, and candidate responses. In the selection layer, the recent parts of the context are used as the query to pre-select the relevant parts of the context $c$ and knowledge collection $k$.
In the matching layer, the context $c$ and knowledge collection $k$ interacts with the candidate response $r$ respectively by the attention mechanism.
In the aggregation layer, another BiLSTM is adopted to extract matching signals considering the dependency among utterances in the context.
As shown in table \ref{tab:example}, sentences in the knowledge collection might be independent of each other. Therefore, matching information of the knowledge collection is aggregated by attending to the context and response candidate, which can be regarded as the post-selection.
In the end, the matching feature is fed into the prediction layer to measure the matching degree between the candidate response $r$ and the context-knowledge pair $(
c,k)$.

\subsection{Representation Layer}
We use general pre-trained word embedding, namely GloVe \cite{pennington2014glove} to represent each word of the context, knowledge, and the response candidate. To alleviate the out-of-vocabulary issue, Word2vec \cite{mikolov2013distributed} trained on the task-specific training set and character-level embedding is adopted in the representation layer. We simply concatenate the three embeddings for each word.
Therefore, the representation layer is capable to capture both semantics and morphology of words.      

Formally, given an utterance $u_i$ in a context $c$, a sentence $k_i$ in the knowledge collection and a response candidate $r$, we embed $u_i$, $k_i$, and $r$ as $\mathbf{E}_{u_i} = [\mathbf{e}_{u_{i,1}},\mathbf{e}_{u_{i,2}},...,\mathbf{e}_{u_{i,l_{p_i}}}]$, $\mathbf{E}_{k_i} = [\mathbf{e}_{k_{i,1}},\mathbf{e}_{k_{i,2}},...,\mathbf{e}_{k_{i,l_{q_i}}}]$ and $\mathbf{E}_r = [\mathbf{e}_{r_{1}},\mathbf{e}_{r_{2}},...,\mathbf{e}_{r_{l_r}}]$ 
respectively, where $\mathbf{e}_{u_{i,t}}$, $\mathbf{e}_{k_{i,t}}$ and $\mathbf{e}_{r_{t}}$ are a $d$-dimension vector concatenated corresponding to the $t$-th word in the $u_i$, $k_i$, $r$ respectively, and $l_{p_i}$, $l_{q_i}$, and $l_r$ are the number of words in the sequences respectively. 

\subsection{Encoding Layer}
We employ an BiLSTM to model the bidirectional interactions among words in the utterances, and encode each utterance as a sequence of hidden vectors. The sentence encoding can be denoted as follows,
\begin{align}
u_{i,j} & = \mathtt{BiLSTM}_{1}(\mathbf{E}_{u_i},j),j\in\{1,2,...,l_{p_i}\} \\
k_{i,j} & = \mathtt{BiLSTM}_{1}(\mathbf{E}_{k_i},j),j\in\{1,2,...,l_{q_i}\} \\
r_{j} & = \mathtt{BiLSTM}_{1}(\mathbf{E}_{r},j),j\in\{1,2,...,l_r\}
\end{align}
where $u_{i,j}$ is the encoded representation of the $j$-th word in the $i$-th context utterance,  $k_{i,j}$ is the encoded representation of the $j$-th word in the $i$-th knowledge sentence, and $r_j$ is the encoded representation of the $j$-th word in the response candidate. 
LSTM can be utilized to encode a sequence of input vectors $(w_1,w_2,...,w_n)$ into hidden vectors $(h_1,h_2,...,h_n)$, which can be formulated as follows,
\begin{align}
i_t & = \sigma(U_i w_t + V_i h_{t-1} + b_i) \\
f_t & = \sigma(U_f w_t + V_f h_{t-1} + b_f) \\
o_t & = \sigma(U_o w_t + V_o h_{t-1} + o_f) \\
g_t & = tanh(U_g w_t + V_g h_{t-1} + b_g) \\
C_t & = i_t \circ g_t + f_t \circ C_{t-1} \\
h_t & = o_t \circ tanh(C_t)
\end{align}
where ``$\circ$'' denotes element-wise product.
BiLSTM places one LSTM in the forward direction and another in the backward direction. We concatenates the outputs of the two LSTMs, which is defined as:
\begin{align}
h_{t} = [ \stackrel{\rightarrow}{h_t} ; \stackrel{\leftarrow}{h_t} ]
\end{align}

\subsection{Selection Layer}

\begin{figure}[t!]
    \centering
    \includegraphics[width=0.9\linewidth]{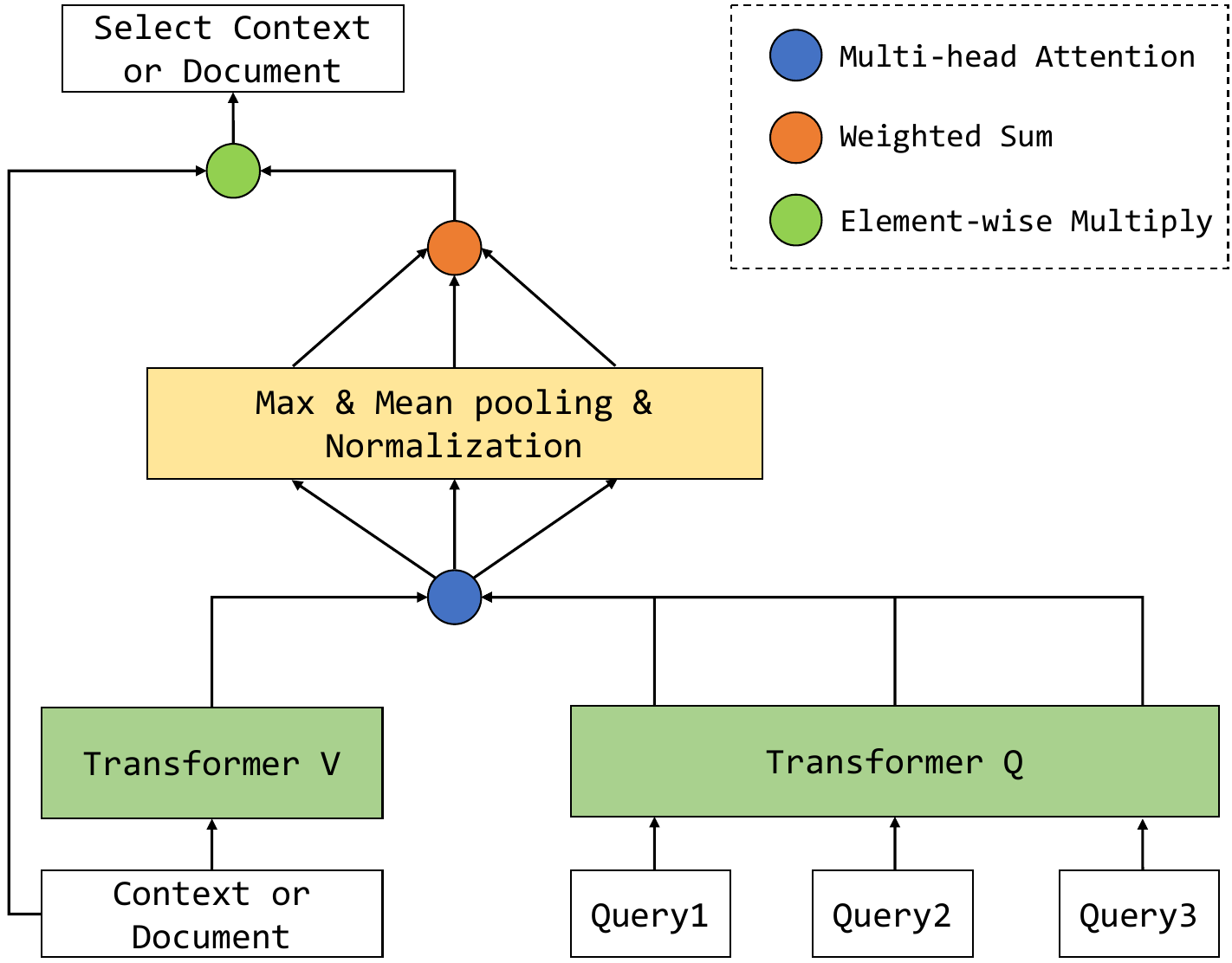}
    \caption{Architecture of the selector. For ease of illustration, we draw three representations for each utterance.}
    \label{fig:selector}
\end{figure}

Figure \ref{fig:selector} depicts the content selector module in RSM-DCK. 
Following the MRFN \cite{tao2019multi}, we use the Attention Module. Three input sentences, namely the query sentence $\mathcal{Q}$, the key sentence $\mathcal{K}$, and the value sentence $\mathcal{V}$, are fed into the Attention Module. To be specific, $\mathcal{Q} = \{e_i\}^{n_q}_{i=1}$, $\mathcal{K} = \{e_i\}^{n_k}_{i=1}$,$\mathcal{V} = \{e_i\}^{n_v}_{i=1}$, where $n_q$,$n_k$, and $n_v$ denote the number of words in each sentence. And $e_i$ is the feature vector of the $i$-th word in the corresponding sequence. The Attention Module uses Scaled Dot-Product Attention \cite{vaswani2017attention} to model the attention mechanism between the $\mathcal{Q}$, $\mathcal{K}$, and $\mathcal{V}$, which can be defined as:
\begin{equation}
    \mathtt{Attention}(\mathcal{Q},\mathcal{K},\mathcal{V}) = \text{softmax}(\frac{\mathcal{Q}  \mathcal{K}^{T}} {\sqrt{d}}) \mathcal{V} \label{eq:4}
\end{equation}
Intuitively, each word of $\mathcal{V}$ is weighted by an importance score calculated by the similarity between a word in $\mathcal{Q}$ and a word in $\mathcal{K}$. By equation \ref{eq:4}, we can obtain $\mathcal{O}$, a new representation of $\mathcal{Q}$, in which the new feature vector of each word in $\mathcal{Q}$ is a weighted sum of the representation of words in $\mathcal{V}$. To improve the expressiveness of the module, a linear transformation operation is adopted before the attention operation. A layer normalization operation \cite{ba2016layer} is applied to $\mathcal{O}$, which helps ease vanishing or exploding of gradients. Then a feed-forward network FFN with RELU is applied upon the normalization result, which helps further enhance the expressiveness of the module. A residual connection is adopted to help gradient flow. Finally, the multi-head mechanism is applied to help the attention module capture diverse features of the input.

In RSM-DCK, the last utterance of the context is used as a query $q$ (namely $u_{l_{p_i}}$) for the reason that the last utterance is usually the most related utterance with the response.


\subsubsection{Context selector}
 To capture long-term dependency in the sequences, we use self-attention layer to further transform the query $q$ and each utterance $u_i$ in the context, which can be formulated as follows,
\begin{align}
    \hat{Q} & = \mathtt{Attention}(Q,Q,Q) \\
    \hat{U}_i & = \mathtt{Attention}(U_i,U_i,U_i) 
\end{align}
where $Q \in \mathbb{R}^{l_{p_i} \times 2d}$ is the representation matrix of $u_{n_c}$ produced by the encoding layer with $l_q$ is the number of words in the query sentence; $U_i \in \mathbb{R}^{l_{p_i} \times 2d}$ denotes the representation matrix of $u_{i}$ and $k_{i}$ produced by the encoding layer.

To pre-select the relevant parts of the context, we use the scaled dot-product attention to calculate the similarity matrix between the query utterance and the previous utterances in the context, which is defined as:
\begin{equation}
    \phi_i = \frac{\hat{Q}\cdot {\hat{U}_{i}}^{T}}{\sqrt{2d}}
\end{equation}
where $\phi_i \in \mathbb{R}^{l_q \times l_{p_i}}$. The operation is applied to each utterance in the context, thus we obtain $\Phi \in \mathbb{R}^{l_q \times n_c \times l_{p_i}}$, where $n_c$ is the number of utterances in the context. From the perspective of the query, a max pooling operation is adopted to select the most related context:
\begin{equation}
    \Tilde{\Phi} = \max \limits_{dim = 0} \Phi
\end{equation}
where $ \Tilde{\Phi} \in \mathbb{R}^{n_c \times l_{p_i}}$

To obtain the relevant score between the query and each utterance in the context, we perform max pooling and mean pooling operation over $\Tilde{\Phi}$, so as to summarizes the maximum value and average value of the attention weight, formulated as:
\begin{align}
    s_1 &= \max \limits_{dim = 1}\Tilde{\Phi}, \\
    s_2 &= \mathop{mean} \limits_{dim = 1} \Tilde{\Phi}
\end{align}
where $s_1,s_2 \in \mathbb{R}^{n_c}$. We then use a hype-parameter $\alpha$ to fuse the two scores:
\begin{align}
   s & = \alpha \cdot s_1 + (1- \alpha) \cdot s_2 \\
    s & = \mathtt{softmax}(s) 
\end{align}
We set initial value of $\alpha$ to $0.5$ and make it update during the training.

Following the MSN \cite{yuan-etal-2019-multi}, we also perform multi-hop selections to enhance the robustness of the content selection module. 
We first concatenate the representation of last $m$ utterances in the context $\{u_{n_c-m+1}, u_{n_c-m+2},\cdots, u_{n_c}\}$, and then carry out a mean-pooling operation to obtain the new query representation $Q^{(m)}$.  We use the new query $q$ to conduct the above selection process, thus obtain $m$ different weight distributions $s_c = [s^{(1)},s^{(2)},\cdots,s^{(m)}]$. To fuse the weight distributions across different hops, we utilize an adaptive combination to compute the final weight distribution $\bar{s}_{c} \in \mathbb{R}^{n_c \times 1}$ as follows,
\begin{align}
   \bar{s}_c & = s_c \cdot \pi
\end{align}
where $\pi \in \mathbb{R}^{k \times 1 }$ is a learnt weight.

Finally, we can obtain the updated representation $\bar{U}_i$ for $i$-th utterance ($u_i$) in the context, which can be formulated as follows,
\begin{equation}
   \bar{U}_i = {U}_i + \bar{s}_c(i) * {U}_i 
\end{equation} 
where $\bar{s}_c(i)$ is the relevance score for $i$-th utterance. The the residual connection is used here to prevent the gradient vanishing. 

\subsubsection{Knowledge selector}
Similarly, we also employ a self-attention layer to process each utterance $k_i$ in the knowledge, which can be formulated as follows,
\begin{align}
    \hat{K}_i & = \mathtt{Attention}(K_i,K_i,K_i)
\end{align}
where $K_i \in \mathbb{R}^{l_{q_i} \times 2d}$ represents the representation matrix of  $k_{i}$ produced by the encoding layer.

Then we can calculate the final weight distribution $\bar{s}_k$ and the updated knowledge representation $\bar{K}_i$ by replacing the $\hat{U}_i$ with $\hat{K}_i$ in the formulation of context selector.
\begin{equation}
   \bar{K}_i = {K}_i + \bar{s}_k(i) * {K}_i 
\end{equation}

To conclude, we use the latter parts of the context to pre-select relevant parts of the context and knowledge collection in the selection layer, which helps RSM-DCK focus on the relevant contents in the following matching and aggregation stage. In addition, RSM-DCK is capable of obtaining the explicit weight distribution indicating the relative importance of each utterance in the context or each sentence in the knowledge collection. 

\subsection{Matching Layer}
Following the ESIM \cite{chen-etal-2017-enhanced}, we use the cross attention mechanism to model
 the interaction between  context and the candidate response, and the interaction between the knowledge collection and the candidate response. The context with multi-turn utterances is concatenated as a long sequence, which can be formally expressed as $C=[U_1,U_2, \cdots, U_{n_{c}}] \in \mathbb{R}^{l_c \times 2d} $ with $l_c$ the total number of words in the context (namely $l_c = \sum_{i=1}^{n_c} l_{p_i}$ ). We then use dot product to calculate the similarity between the $i$-th word in the context $c$ and the $j$-th word in the candidate response $r$, which can be formulated as follows,
 \begin{equation}
 \mathcal{E} = C^T \cdot R  
 \end{equation}
where  $R \in \mathbb{R}^{l_{r} \times 2d}$ denotes the representation matrix of $r$ produced by the encoding layer; $ \mathcal{E}  \in \mathbb{R}^{l_c \times l_r}$  is the attention matrix. We can obtain the weight matrices $\mathbf{\alpha} \in \mathbb{R}^{l_c \times l_r}$ and $\mathbf{\beta} \in \mathbb{R}^{l_c \times l_r}$ by normalization alongside the column and the row respectively, which can be formulated as follows, 
 \begin{align}
  \alpha_{i,j} &= \frac{\exp( \mathcal{E}_{i,j})}{\sum_{j=1}^{l_r} \exp( \mathcal{E}_{i,j})} \\
  \beta_{i,j} &= \frac{\exp(\mathcal{E}_{i,j})}{\sum_{i=1}^{l_c} \exp( \mathcal{E}_{i,j})}
 \end{align}

Through a weighted combination, we can obtain  a response-aware context representation $\hat{C}$ and a context-aware response representation $\hat{R}$ for the context and the response candidate respectively, formulated as: 
  \begin{align}
     \hat{C} & = \alpha \cdot R \\
     \hat{R} & = C \cdot  \beta 
 \end{align}

  Referring to TBCNN
  \cite{mou-etal-2016-natural}, to measure the aligned token-level semantics similarity, 
  we introduce the following functions to obtain the matching features $\bar{C}$ and $\bar{R}$,
  \begin{align}
  \bar{C} &= [C,\hat{C} ,C-\hat{C}, C \circ \hat{C}] \\
   \bar{R} &= [R,\hat{R},R-\hat{R},R \circ \hat{R}]
  \end{align}
  where ``$\circ$'' denotes element-wise product operation. 

    
Similarly, we also concatenated all knowledge entries as a long sequence $K^c$, and then perform the above cross attention, thus obtaining a response-aware knowledge representation $\hat{K}^c$ and a knowledge-aware response representation $\hat{R}^c$ respectively. The two representations are further transformed into the matching features $\bar{K}^c$ and $\bar{R}^c$ via the following formulations,
  \begin{align}
  \bar{K}^c &= [K^c,\hat{K}^c ,K^c-\hat{K}^c, K^c \circ \hat{K}^c] \\
   \bar{R}^c &= [R,\hat{R}^c,R-\hat{R^c},R \circ \hat{R}^c]
  \end{align}

\begin{table*}[ht!]
\centering
\caption{Statistics of the two datasets.}
\resizebox{0.6\textwidth}{!}{
\begin{tabular}{|l|c|c|c|c|c|c|} 
\hline
\multirow{2}{*}{Statistics}
& \multicolumn{3}{c|}{Persona-Chat} 
& \multicolumn{3}{c|}{CMUDoG} 
\\ \cline{2-7}
& Train  & Val  & Test 
& Train  & Val  & Test  
\\ \hline
\# of conversations  & 8939   & 1000  & 968  & 2881  & 196   & 537 \\ \hline
\# of turns    & 65719  & 7801  & 7512 & 36159 & 2425  & 6637 \\ \hline
Av\_turns / conversation   & 7.35   & 7.80  & 7.76 & 12.55 & 12.37  & 12.36 \\ \hline
Av\_length of utterance     & 11.67 & 11.94 & 11.79 & 18.64 & 20.06 & 18.11\\ \hline
\end{tabular}
}
\label{tbl:stat}
\end{table*}

\subsection{Aggregation Layer}
In the aggregation layer, matching information is first aggregated on the sentence-level. We convert the representation of the context $ \bar{C} $ back to the  representations of each utterance as $\{\ddot{U}_i\}^{n_c}_{i=1}$.  Then, an BiLSTM is adopted, followed by max pooling and last hidden state pooling to obtain a fixed-length vector for each utterance in the context. The process can be formally expressed as follows,
\begin{align}
\label{con:9}
\dddot{U}_{i,j} & =  \mathtt{BiLSTM}_{2}(\ddot{U}_i,j),j\in\{1,2, \cdots ,l_{p_i}\} 
\\
 U_{i}^s &= [ \mathtt{max}(\dddot{U}_i); \dddot{U}_{i,l_{p_i}} ]
\label{con:10}
\end{align}
where $ U_{i}^s$ denotes the matching feature between the $i$-th utterance and the response candidate. We also process the context-aware response representation $\bar{R}$ and the knowledge-aware response representation $\bar{R}^c$ with the same operation described in Equation~(\ref{con:9}-\ref{con:10}),  and obtain aggregated features $M_r$ and $M_r^c$ respectively.

To aggregate the matching information in the session-level, we adopt another BiLSTM (denoted as $\mathtt{BiLSTM}_3$) to process a sequence of matching feature $\{U_{i}^s\}_{i=1}^{n_c}$, and take the concatenation of the final hidden vector and the max-pooling of the hidden vectors as the final matching feature between the context and the response candidate, denoted as $M_{c}$.

For the knowledge $ \bar{K}^c $, similarly, we first convert the representation into the representations of each sentence $\{\ddot{K}^c_i\}_{i=1}^{n_k}$. Then we process each sentence in the knowledge collection with the formulation in Equation~(\ref{con:9}-\ref{con:10}), and obtain the matching feature between $i$-th sentence in the knowledge collection and the response candidate, denoted as $ K_{i}^s$.
Different from the diglogue context, sentences in the knowledge collection are relative independent to each other, so we aggregate the matching information of the knowledge collection via the attention mechanism, which can be regarded as the post-selection for knowledge. To be specific, we use $M^{c}_{r}$ which contains high-level information about the dialogue context and candidate response to attend to $K^{s}$, thus obtaining $M_{k}$. The process is defined as:
\begin{align}
\gamma_i &= \frac{\exp( K^{s}_{i} \cdot M^{c}_{r} )}{\sum_{i=1}^{n_k} \exp( K^{s}_{i} \cdot M^{c}_{r})}\label{eq:27} \\
M_k &= \sum_{i=1}^{n_k}  \gamma_{i}  K^{s}_{i}\label{eq:28}
\end{align}

Finally, we take the concatenation of $M_c$, $M_k$, $M_r$, and $M_r^c$ as the final matching feature, formulated as:
\begin{equation}
M_\mathtt{final} = [M_c;M_k;M_r;M_r^c]
\end{equation}

\subsection{Prediction Layer}
The final matching feature vector $M_\mathtt{final}$ is then fed into a multi-layer perceptron (MLP) classifier to get the matching score between the context, the knowledge and the candidate response, which is defined as:
\begin{equation}
g(c,k,r) =f_2 (W_2^{\top} \cdot f_1( W_1^{\top} \cdot {M}_\mathtt{final} + b_1) + b_2), \label{eq:pred}
\end{equation}
where $ W_{1},  W_{2}, b_{1}$, and $ b_{2} $ are parameters. $f_1(\cdot)$ is the tanh activation function, and $f_2(\cdot)$ is the softmax function.

\subsection{Learning Method}
As is shown in Equation (\ref{eq:pred}), we adopt a softmax normalization layer over the matching logits of all candidate responses to speed up training and alleviate the imbalance of samples. Parameters of the model are optimized by minimizing the cross-entropy loss on $D$, which is defined as
\begin{equation}
\mathcal{L}({\Theta}) = -\sum_{i=1}^{N} y_{i}  \log g(c_i,k_i,r_{i})
\end{equation}
where $\Theta$ represents the parameters of the model.

\section{Experiments}

\subsection{Datasets and Evaluation}
We test RSM-DCK on two benchmark datasets, namely the Persona-Chat dataset \cite{zhang2018personalizing} and the CMUDoG dataset \cite{cmu_dog_emnlp18}.
Statistics of the two datasets are shown in table \ref{tbl:stat}.

\paragraph{\textbf{Persona-Chat}}
The first dataset we use is the Persona-Chat \cite{zhang2018personalizing}, which consists of 151,157 turns of conversations. The dataset is split as a training set, a validation set and a testing set by the publishers. In all the three sets, each utterance is associated with one positive response comes from the groud-truth and 19 negative response candidates that are randomly sampled from the dataset.  
To create the dataset, crowd workers are randomly paired and each of them is required to chat naturally with his partner according to his assigned profiles.
The persona profiles describe the characteristics of the speaker, thus can be regarded as a collection of knowledge which consists of 4.49 sentences on average.
For each dialogue, there are two format personas, namely, original profiles and revised profiles. The latter is rephrased from the former, which helps prevent models utilizing word overlap and makes response selection more challenging.

\paragraph{\textbf{CMUDoG}}
Another dataset we use is the CMUDoG dataset published in \cite{cmu_dog_emnlp18}. The conversations about some certain documents are collected through Amazon Mechanical Turk. The topic of the documents is all about movies, thus helping interlocutors have a common topic to talk about naturally. To impel two paired workers to talk about the given documents, two scenarios are explored. In the first one, only one interlocutor can see the document while the other cannot. The interlocutor who has access to the given document is instructed to introduce the movie to the other. In the second one, both interlocutors have access to the given document and are required to talk about the contents of the document. Considering the small number of conversations in each scenario, data in the two scenarios is merged to form a larger dataset.  Following the DGMN \cite{ijcai2019-756}, we filter out conversations whose turns are less than 4 to avoid noise. Due to the lack of negative responses in the dataset, we sample 19 negative response candidates for each utterance from the same set.

Following previous studies~\cite{cmu_dog_emnlp18,ijcai2019-756}, we also employ $R_n@k$ as evaluation metrics, where $R_n@k$ ($n$=20, $k$={1,2,5}) stands for recall at position $k$ in $n$ response candidates.


\begin{table*}[t!]
\centering
\caption{Evaluation results on the test sets of the Persona-Chat data and the CMUDoG data. Numbers in bold mean that improvement over the best baseline is statistically significant (t-test, $p$-value $<0.01$).}
\vspace{-2mm}
\resizebox{0.85\textwidth}{!}{
\begin{tabular}{|l|c|c|c|c|c|c|c|c|c|}
\hline
\multirow{2}{*}{\backslashbox{Models}{Metrics}}  
&  \multicolumn{3}{c|}{Original Persona}  
& \multicolumn{3}{c|}{Revised Persona}  
& \multicolumn{3}{c|}{CMUDoG} 
\\ 
\cline{2-10}
        & R$_{20}@1$ & R$_{20}@2$ & R$_{20}@5$ & R$_{20}@1$ & R$_{20}@2$ & R$_{20}@5$  
        & R$_{20}@1$ & R$_{20}@2$ & R$_{20}@5$ \\ \hline
IR Baseline~\cite{zhang2018personalizing}     & 41.0  & -     & -     & 20.7     & -     & -    & -     & -     & -   \\ 
Starspace~\cite{wu2018starspace}       & 49.1  & 60.2     & 76.5     & 32.2     & 48.3     & 66.7   & 50.7     & 64.5     & 80.3  \\ 
Profile Memory~\cite{zhang2018personalizing}  & 50.9  & 60.7     & 75.7     & 35.4     & 48.3     & 67.5  & 51.6  & 65.8    & 81.4   \\
KV Profile Memory~\cite{zhang2018personalizing}  & 51.1 & 61.8    & 77.4     & 35.1     & 45.7     & 66.3  & 56.1     & 69.9     & 82.4 \\ 
Transformer~\cite{mazare2018training} & 54.2 & 68.3    & 83.8     & 42.1     & 56.5     & 75.0  & 60.3    & 74.4     & 87.4 \\ 
DGMN~\cite{ijcai2019-756}             & 67.6 & {80.2} & {92.9}  & {58.8} & {62.5} & {87.7}  & {65.6} & {78.3} & {91.2} \\ 
DIM  ~\cite{gu-etal-2019-dually}           & 78.8 & - & -  & {70.7} & - & -  & {78.58} & {88.41} & {96.47} \\ \hline
RSM-DCK & \textbf{79.65}    & \textbf{90.21}     & \textbf{97.47}    &\textbf{71.85}   & \textbf{84.94} & \textbf{95.50}   & \textbf{79.25}     & \textbf{88.84}     & \textbf{96.66}     \\ \hline
\end{tabular}
}
\label{tab:main_results}
\end{table*}

\subsection{Baseline Models}
We compare our model with the following models:
\paragraph{\textbf{IR Baseline} \cite{zhang2018personalizing}}
An information retrieval method which selects the appropriate response based on simple word overlap.

\paragraph{\textbf{Starspace} \cite{wu2018starspace}} The model learns the task-specific embedding by minimizing the margin ranking loss and returns
the cosine similarity between the conversation context concatenated by the associated document and the response candidate.

\paragraph{\textbf{Profile Memory} \cite{zhang2018personalizing}} The model uses the memory network as the backbone. To be specific, the model uses the dialogue history as the query, and performs attention over the profile sentence to find the relevant lines to be combined with the query, and then measures the similarity between the fused query and response candidate.

\paragraph{\textbf{KV Profile Memory} \cite{zhang2018personalizing} } The model extends the Profile Memory to a multi-hop version.
In the first hop, the model uses the Profile Memory to obtain the fused query. Then, in the second hop, the attention is conducted with the dialogue history as the key and the next dialogue utterance as the value. The output of the second hop is utilized to measure the similarity between itself and the response candidate.

\paragraph{\textbf{Transformer} \cite{mazare2018training} }
The model encodes the dialogue history and response candidate with Transformer encoder \cite{vaswani2017attention}, and encodes the profile sentences via the bag-of-words representation instead.  

\paragraph{\textbf{DGMN}\cite{ijcai2019-756}} The model encodes representations of the dialogue context and document with the self-attention, and fuses representations of the dialogue context and document into each other by the cross-attention, then interacts with the response candidate via the hierarchical attention individually.

\paragraph{\textbf{DIM}\cite{gu-etal-2019-dually}} The model encodes representations of the dialogue context, document, and response candidate with the BiLSTM and lets the dialogue context and document interact with the response candidate respectively via the cross-attention mechanism. Finally, another BiLSTM is adopted to aggregate the matching features of the dialogue context, document, and response candidate. This model outperforms all baselines above.

\begin{table*}[t!]
\centering
\caption{Ablation study on the Persona-Chat and CMUDoG dataset.}
\vspace{-2mm}
\resizebox{\textwidth}{!}{
\begin{tabular}{|l|c|c|c|c|c|c|c|c|c|}
\hline
\multirow{2}{*}{\backslashbox{Models}{Metrics}}  
&  \multicolumn{3}{c|}{Original Persona}  
& \multicolumn{3}{c|}{Revised Persona}  
& \multicolumn{3}{c|}{CMUDoG} 
\\ 
\cline{2-10}
        & R$_{20}@1$ & R$_{20}@2$ & R$_{20}@5$ 
        & R$_{20}@1$ & R$_{20}@2$ & R$_{20}@5$  
        & R$_{20}@1$ & R$_{20}@2$ & R$_{20}@5$
\\ 
\hline
RSM-DCK & 79.65 & 90.21  & 97.47     &71.85 &84.94  & 95.50  &79.25 &88.84  &96.66   \\ \hline
RSM-DCK (w/o context selector) &78.79 &89.60 &97.16     &71.13 &84.38 &95.15  &77.93 &88.17 &96.53   \\
RSM-DCK (w/o knowledge selector) &78.94 &89.19 &97.00    &71.09 &84.25 &95.49 &78.18 &88.55 &96.61 \\ 
RSM-DCK (w/o knowledge post-selection) &78.87 &89.36 &97.27     &70.79 &84.58 &95.43  &79.19 &89.09 &96.82 \\ \hline
RSM-DCK (w/o context) & 48.38 &  58.21 &  72.95 & 30.10 &  41.12 &  60.90  & 57.01 &  72.13 &  87.99 \\ 
RSM-DCK (w/o knowledge) & 62.94 &  77.37 &  91.57    & 63.03 &  77.64 &  91.00 & 74.21 &  86.02 &  95.45    \\ \hline
\end{tabular}
}
\vspace{-2mm}
\label{tab:ablation}
\end{table*}

\subsection{Implementation Details}
In RSM-DCK, we set the dimension of the glove  embedding as 300, and word2vec is trained on the training set with the dimension of the word embedding set as 100. The dimension of the character-level embedding is set as 150 with window size \{3, 4, 5\}, each containing 50 filters. We freeze the embedding weights during training. The hidden size of all BiLSTMs is set as 300, and the number of head of the Attention Module is 3. The MLP in the prediction layer has 256 hidden units, which is activated by ReLU \cite{nair2010rectified}. In the Persona-Chat dataset, the maximum number of utterances per dialogue context is set as 15, and 5 for sentences per document. We set the maximum number of words as 20 for each utterance in the dialogue context, each sentence in knowledge collection, and each candidate response (i.e., $l_{p_i} = l_{q_i} = l_r = 20$) . If the number of words is less than 20, we pad zeros, otherwise, we keep the latter 20 words. In the CMUDoG dataset, the maximum number of utterances per dialogue context is set as 8, and 20 for sentences per document; $l_{p_i} = l_{q_i} = l_r = 40$.
The Adam method \cite{DBLP:journals/corr/KingmaB14} is applied with learning rate set as 0.00025 for the Persona-Chat dataset with a batch size of 12 and 0.0001 for the CMUDoG dataset with a batch size of 6.

We implement our model with PyTorch and train the model with at most 20 epochs on a 2080ti machine. The early stop is adopted to avoid overfitting. We copy the most results from the DGMN \cite{ijcai2019-756}. For the DIM \cite{gu-etal-2019-dually} on CMUDoG dataset, we use the codes published at \url{https://github.com/JasonForJoy/DIM} to get the results.

\subsection{Evaluation Results}
Table \ref{tab:main_results} shows the evaluation results on the benchmark datasets. We can find that our proposed model outperforms the baseline models in terms of all metrics. Compared with DIM, the previous best performing baseline on all the two benchmarks, our model achieves $0.9\%$ absolute improvement on $R_{20}@1$ on the Original Persona-Chat dataset,  $1.2\%$ absolute improvement on $R_{20}@1$ on the Revised Persona-Chat dataset, and $0.7\%$ absolute improvement on $R_{20}@1$ on the CMUDoG dataset.
{It is worth noting that the improvement on the CMUDoG dataset is relatively smaller than Persona-Chat. The reason might be that the knowledge is more important for Persona-Chat compared with CMUDoG and the model can benefit more from the selection of the knowledge on the Persona-Chat dataset.}

\section{Discussions}
In this section, we investigate the effects of the variance of the dialogue context and knowledge collection on the performance of RSM-DCK. First, we conduct an ablation study to demonstrate the effectiveness of RSM-DCK empirically. Second, we explore how the performance of RSM-DCK varies with respect to the number of utterances in the dialogue context. In addition, we visualize the weights of the context and knowledge collection given by the selection mechanism in RSM-DCK through a case study.

\begin{table*}[t!]
\caption{Performance of models across different numbers of utterances in the context.}
\vspace{-1mm}
\resizebox{0.8\textwidth}{!}{
\begin{tabular}{|l|c|c|c|c|c|c|c|c|c|c|c|c|}
\hline
 & \multicolumn{4}{c|}{Original Persona} & \multicolumn{4}{c|}{Revised Persona} & \multicolumn{4}{c|}{CMUDoG}     \\ \cline{1-13}
Utterances number        & (0,3{]} & (4,7{]} & (8,11{]} & (12,15{]} & (0,3{]} & (4,7{]} & (8,11{]} & (12,15{]} & (0,2{]} & (3,4{]} & (5,6{]} & (7,8{]} \\ \hline
Case Number       & 1936    & 1936      & 1936      & 1704  & 1936    & 1936      & 1936      & 1704  & 537      & 537         & 537          & 5026   \\ \hline
DGMN R$_{20}@1$ & 73.61    & 71.23      & 66.27      & 61.44  & 62.55    & 58.94       & 59.56      & 58.86  & 79.14     & 74.67        &70.20         & 69.34   \\ \hline
DIM R$_{20}@1$  & 82.85    & 81.56      & 76.76      & 71.48  & \textbf{74.33}   & 70.76       & 69.47     & 69.60  & 88.12    & 83.43        &75.98        & 77.32   \\ \hline
RSM-DCK R$_{20}@1$  & $\textbf{84.61}$    & $\textbf{81.66}$      & \textbf{77.65}      & \textbf{74.00}  & 72.56    & $\textbf{71.95}$      & \textbf{72.31}       & \textbf{70.42}  & 87.71      & 83.05         & \textbf{79.52}         & \textbf{77.91}   \\ \hline
\end{tabular}
}
\label{tab:utterances_num}
\end{table*}

\subsection{Ablations}
We perform a series of ablation experiments to investigate the relative importance of the selection mechanism for the context and knowledge. Also, we investigate the relative importance of the context and knowledge collection individually. First, we use the complete model as the baseline, then we conduct ablation experiments as follows:
\begin{itemize}
\item \textbf{w/o context selector}: we remove the dialogue context selector in the selection layer.
\item \textbf{w/o knowledge selector}: we remove the document selector in the selection layer.
\item \textbf{w/o knowledge post-selection}: we remove the post-selection for knowledge in the aggregation layer, and replaced it with a simple mean pooling operation.
\item \textbf{w/o context}: we remove the dialogue context from the model. The setting means that the response is selected according to the knowledge collection only.
\item \textbf{w/o knowledge}: we remove the knowledge collection from the model. The model becomes a traditional context-response matching architecture.
\end{itemize}

\begin{figure}[t!]
\centering
\resizebox{\columnwidth}{!}{
\begin{tabular}{|c|l|} 
\hline  
$\bar{s}_c$ & \makecell[c]{Context}    \\ \cline{1-2}
 \cellcolor[rgb]{0.80632065, 0.92189158, 0.61790081}0.147  & hello what are doing today? \\ 
 \cellcolor[rgb]{1., 1., 0.89803922}0.092  & 
 i am good, i just got off work and tired, i have two jobs. \\ 
 \cellcolor[rgb]{0.49341023, 0.78637447, 0.48355248}0.186  & 
 i just got done watching a horror movie \\ 
 \cellcolor[rgb]{0.89434833, 0.95852364, 0.67101884}0.131  & i rather read, i have read about 20 books this year. \\ 
 \cellcolor[rgb]{0.86851211, 0.94818916, 0.65207228}0.137  & wow! i do love a good horror movie. loving this cooler weather \\ 
 \cellcolor[rgb]{0.56513649, 0.81750096, 0.51197232}0.178  & but a good movie is always good. \\ 
 \cellcolor[rgb]{0.89803922, 0.96, 0.67372549}0.130  & yes! my son is in junior high and i just started letting him watch them too  \\ \hline
True response & i work in the movies as well.    \\ \cline{1-2}
False response & that is great ! are you going to college ? \\ \cline{1-2}
\end{tabular}
}
\caption{Weights of each utterance in the context.}

\label{tab:vis1}
\end{figure}

\begin{figure}[t!]
\centering
\resizebox{0.8\columnwidth}{!}{
\begin{tabular}{|c|l|c|}\hline 
$\bar{s}_k$ &\makecell[c]{Knowledge collection}   & $\gamma$    \\ \cline{1-3}
\cellcolor[rgb]{0.6042599, 0.83447905, 0.52747405}0.278  & i read twenty books a year. & 0.0006 \\ 
\cellcolor[rgb]{0.87958478, 0.95261822, 0.66019223}0.206  & i am a stunt double as my second job. 
& \cellcolor[rgb]{0.11680123, 0.50128412, 0.25573241}0.9987\\ 
\cellcolor[rgb]{0.16032295, 0.54763552, 0.28273741}0.390  & i only eat kosher. & 0.0001\\ 
\cellcolor[rgb]{1., 1., 0.89803922}0.126  & i was raised in a single parent household. & 0.0005\\ 
\hline
\end{tabular}
}
\vspace{-1mm}
\caption{Weights of each entry in the knowledge collection.}
\vspace{-2mm}
\label{tab:vis2}
\end{figure}

Table \ref{tab:ablation} shows the evaluation results of the ablation studies. We can observe that removing any of the context selector, knowledge selector, and knowledge post-selection leads to the performance drop compared with the complete model, which demonstrates the effectiveness and necessity of each component of RSM-DCK. 
Besides, we find that the context selector and knowledge selector show a comparable role on the Persona-Chat dataset, while the context selector is generally more useful than the knowledge selector on the CMUDoG dataset as the performance of the model drops more when the context selector is removed.
According to the results of the last two rows in Table~\ref{tab:ablation}, we can observe that removing the dialogue context leads to more degradation of performance of models than removing the knowledge collection, indicating that the dialogue context is more important than the knowledge collection for response selection. 
It should be noticed that the performance of RSM-DCK on the Persona-Chat dataset drops more dramatically than that on the CMUDoG dataset when the knowledge collection is removed from the matching model. The result implies that knowledge collection is more important for recognizing the proper response candidate on the Persona-Chat dataset.

\subsection{Length Analysis}
We further study how the number of utterances in the dialogue context influences the performance of RSM-DCK by binning the test samples into different buckets according to the number of utterances in the dialogue context. To be noticed, the maximum number of utterances per dialogue context is set as 8 in the CMUDoG dataset, which is different from 4 and 15 set by DGMN and DIM respectively. Thus, we run the codes of DGMN and DIM under our settings.
As shown in Table \ref{tab:utterances_num}, the performance degrades as the number of utterances rises up. The reason is that the topic shifting exists in the dialogue context, and the model may suffer from the new topic unrelated with the dialogue history as the number of utterances rises up, thus leading to performance decrease. 
Besides, we can also observe that RSM-DCK shows a stronger capability of handling the dialogue context with more utterances than other models. The main reason is that RSM-DCK can select the relevant parts of the dialogue context and thus achieve better understanding of the dialogue context, with the help of the context selector in the selection layer.

\subsection{Case Study}
To further understand how RSM-DCK performs content selection for the dialogue context and knowledge collection, we visualize the relevance score (attention weight) for each entry of the knowledge or the context.
Figure \ref{tab:vis1} shows the relevance scores of each utterance (a.k.a. $\bar{s}_c$) in the dialogue context. Figure \ref{tab:vis2} shows the pre-selection scores (a.k.a. $\bar{s}_k$) and the post-selection scores (a.k.a. $\gamma$) for each entry in the knowledge collection at the left side and right side respectively. 
As shown in Figure \ref{tab:vis1}, the latter parts of the dialogue context, with the key word ``movie" and key phrase ``horror movie", 
indicate that the topic of the dialogue might be related to movies. Therefore, RSM-DCK rates highly for the third and sixth utterances which are strongly relevant to movies. The result shows that RSM-DCK can properly detect the relevant parts of the dialogue context. As shown in Figure \ref{tab:vis2}, the second sentence in the knowledge collection should be utilized for choosing the true response. Though the pre-selection for the knowledge collection fails, the post-selection still gives the second sentence the highest score, which helps the model out in the dilemma of deciding which response candidate is more sensible.

\section{Conclusion and Future Work}
In this paper, to overcome the performance drop caused by irrelevant parts contents existing in the dialogue context and document, we propose a multi-turn \textbf{R}esponse \textbf{S}election \textbf{M}odel with carefully designed selection mechanism that can \textbf{D}etect the relevant parts of the \textbf{C}ontext and \textbf{K}nowledge collection (\textbf{RSM-DCK}). We conduct experiments on two benchmark datasets and our model achieves new state-of-the-art performance, which shows the superiority of our model. 
In the future, we would like to apply the content selection mechanism to KG-grounded dialogue systems to select the relevant triples in the knowledge graph and generation-based conversation. Furthermore, we also plan to combine our model with the pre-trained model (such as Bert \cite{devlin-etal-2019-bert}) to further improve the performance of response selection.

\begin{acks}
We would like to thank the anonymous reviewers for their constructive comments and valuable suggestions. This work was supported by the National Key Research and Development Program of China (No. 2017YFC0804001), the National Science Foundation of China (NSFC Nos. 61672058 and 61876196). Rui Yan is partially supported as a Young Fellow of Beijing Institute of Artificial Intelligence (BAAI).
\end{acks}

\bibliographystyle{acmart}
\bibliography{acmart}

\end{document}